\ifdefined\paperversion\else
  \def\paperversion{arxiv}%
\fi
\def\versionicra{icra}
\def\versionarxiv{arxiv}
\newif\ifarxiv
\ifx\paperversion\versionarxiv
  \arxivtrue
\else\ifx\paperversion\versionicra
  \arxivfalse
\else
  \errmessage{Unknown \string\paperversion\space `\paperversion'; use icra or arxiv}
\fi\fi

\ifdefined\pdfminorversion\pdfminorversion=4\fi
\documentclass[letterpaper, 10 pt, conference]{ieeeconf}

\IEEEoverridecommandlockouts
\usepackage{graphicx}
\usepackage{amsmath,amssymb}
\usepackage{xcolor}
\usepackage[hidelinks]{hyperref}
\usepackage{algorithm}
\usepackage{algpseudocode}
\ifarxiv
\usepackage{array}  %
\fi

\hypersetup{
    pdfauthor={Maitree Meher, Siming He, S. Shankar Sastry,
               Jitendra Malik, Claire Tomlin, Jonas Frey}
}

\makeatletter
\renewcommand{\@IEEEfigurecaptionsepspace}{\vskip -12pt\relax}
\makeatother

\title{\LARGE \bf
Racing the Clock: Teaching VLNs Budget-Conditioned Object Search
}

\author{
Maitree Meher$^{1*}$,
Siming He$^{1*}$,
S.~Shankar Sastry$^{1}$,
Jitendra Malik$^{1}$,
Claire Tomlin$^{1}$,
Jonas Frey$^{1,2}$
\thanks{$^{1}$University of California, Berkeley
\qquad $^{2}$Stanford University}
\thanks{$^{*}$Equal contribution.}
}

\begin{document}

\maketitle
\thispagestyle{empty}
\pagestyle{empty}

\begin{abstract}
\ifarxiv
How do action budgets shape vision-language navigation (VLN) exploration, and what expert supervision best teaches budget-aware search? We study \emph{budget-conditioned object search}: finding as many distinct instances of a requested category as possible within a budget, with discovery defined geometrically by camera field of view and proximity. Distinct discoveries provide a denser evaluation signal than single-target success. We cast expert design as a choice of \emph{scene abstraction}: a single budget-indexed Bellman recursion, applied to different abstractions of the full environment, yields experts that differ in what they know. An \emph{object-level Bellman expert} is given every target location and reduces search to budgeted routing. A \emph{room-level Bellman expert} retains observed room connectivity, doorway evidence, and cross-scene semantic priors but withholds the contents of unexplored rooms, so it must search for targets that its student also cannot see. We distill both experts into Uni-NaVid under matched training conditions and also evaluate greedy-frontier supervision on held-out HM3D scenes. Changing only the stated budget changes learned trajectories. Despite higher object-level expert return, room-level supervision achieves the highest mean collection among evaluated VLNs at every tested budget. At 2000 steps, it finds 72.7\% of reachable targets, versus 68.5\% under greedy-frontier supervision, 65.5\% under object-level supervision, and 44.8\% for the pretrained VLN. These results show that expert-return rankings need not predict supervision quality for VLN exploration, and motivate choosing an expert's abstraction with the student's information in mind rather than by expert return alone.
\else
How do action budgets shape vision-language navigation (VLN) exploration, and how does the source of supervision affect learned search? We study \emph{budget-conditioned object search}: finding as many distinct instances of a requested category as possible within a budget, with discovery defined geometrically by camera field of view and proximity. Distinct discoveries provide a denser evaluation signal than single-target success. We compare two supervision schemes: an \emph{object-level Bellman expert} routes among targets with the full map and locations, while a \emph{room-level Bellman expert} represents unexplored rooms through semantic estimates and cross-scene priors, leaving their unseen contents unknown. Both optimize budgeted Bellman objectives within their scene abstractions. We distill them into Uni-NaVid under matched training conditions and also evaluate greedy-frontier supervision on held-out HM3D scenes. Changing only the stated budget changes learned trajectories. Despite higher object-level expert return, room-level supervision achieves the highest mean collection among evaluated VLNs at every tested budget. At 2000 steps, it finds 72.7\% of reachable targets, versus 68.5\% under greedy-frontier supervision, 65.5\% under object-level supervision, and 44.8\% for the pretrained VLN. These results show that expert-return rankings need not predict supervision quality for VLN exploration and motivate accounting for budget and expert search behavior when collecting demonstrations.
\fi
\end{abstract}
 
\section{Introduction}
\begin{figure*}[!t]
\centering
\includegraphics[width=\textwidth]{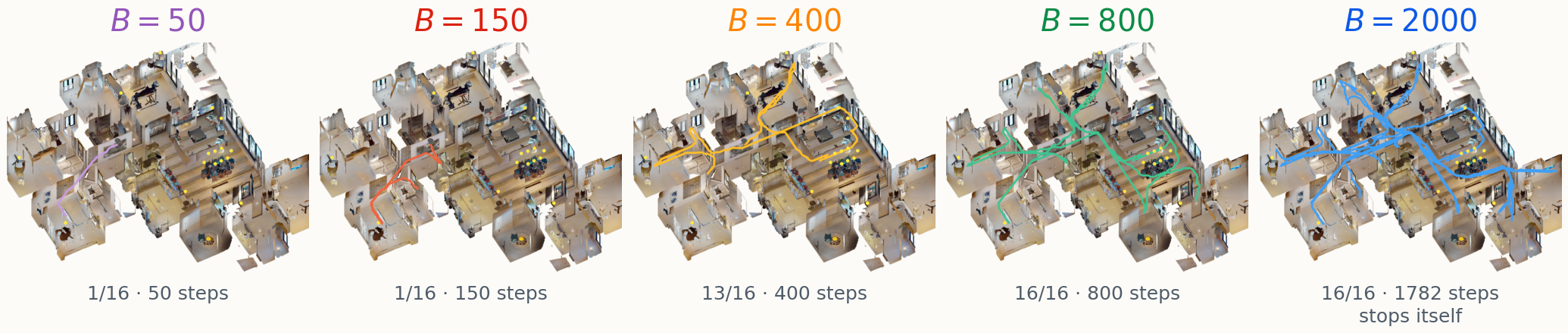}
\caption{\textbf{The stated budget changes the learned VLN's search strategy.}
Room-level Bellman-supervised Uni-NaVid is rolled out from the same start pose in the
same held-out HM3D scene with identical deterministic decoding; only the
budget in the instruction changes. The resulting trajectories differ
substantially rather than simply extending one another, most visibly between
$B=150$ and $B=400$. Across $B=50/150/400/800/2000$, the policy collects
$1/1/13/16/16$ of 16 reachable chairs.}
\label{fig:teaser}
\end{figure*}

An indoor search robot must allocate its budget between nearby targets and distant regions: a short budget may favor immediate collection, while a longer one may justify travel to a promising room. To assess both effective budget use and its influence on decisions, we first ask: \emph{how does the available budget change where and how a vision-language navigation (VLN) policy searches?}

This motivates \emph{budget-conditioned object search}: finding as many distinct instances of a requested category as possible within an action budget. The agent receives RGB history, a target instruction, and a remaining-budget countdown, but no complete map or unseen target locations. Travel, scanning, and target approach consume budget. We count an instance as \emph{collected} when it meets geometric proximity and camera-field-of-view criteria, without physical pickup. Distinct discoveries provide a denser signal than binary success and preserve repeated choices between visible targets and further exploration. Collection across budgets measures search performance; controlled countdown interventions test behavioral sensitivity (Fig.~\ref{fig:teaser}).

The remaining horizon can change the optimal action despite identical observation histories~\cite{pardo2018timelimits}; omitting it creates \emph{state aliasing} between short-horizon collection and longer-horizon exploration. Although finite-resource planning is well studied, pretrained VLNs such as Uni-NaVid~\cite{uninavid} lack an explicit remaining-budget countdown. In a NaVILA~\cite{navila} pilot, a shorter stated limit sometimes produced more actions than a longer one, motivating budget-aware demonstrations beyond merely adding a prompt limit.

To assess what supervision best transfers to a partially observed student, we next ask:
\emph{what kind of expert demonstrations effectively teach budget-conditioned search?} A full-map expert can route efficiently yet prescribe different actions in environments indistinguishable to the student, an information mismatch that can hinder asymmetric imitation~\cite{NEURIPS2021_9fc66491,pmlr-v139-warrington21a,kim2025distillingrealizablestudentsunrealizable,ICLR2026_398753dd}.
\ifarxiv
Prior work addresses this mismatch by modifying the imitation objective or by adapting the expert to the student. We ask a complementary question: \emph{what environmental information should the planning expert itself retain when generating exploration supervision?}

Existing exploration methods span different information regimes. Experts with full map and target-location access~\cite{spoc,header} reduce search to routing among known goals, while receding-horizon planners~\cite{activenerf,bircher2016receding,11128798,baek2025pipe} act from currently available information but may provide weak long-horizon guidance. Scene-prediction methods inject priors about unseen regions~\cite{10610464,harutyunyan2025mapexrl}, but predicting detailed unseen geometry and semantics is itself challenging.

We make this information choice explicit. An expert plans over an abstraction $z=\varphi(\mathcal{M})$ of the full environment $\mathcal{M}$ and knows only the part of $z$ revealed to it so far; one budget-indexed Bellman recursion over this information state then defines the expert. Varying $\varphi$ spans a spectrum of information access. Revealing every target location yields an \emph{object-level Bellman expert} that routes among known targets: nothing is left to search for. Retaining observed room connectivity, doorway evidence, and cross-scene semantic priors while withholding the contents of unexplored rooms yields a \emph{room-level Bellman expert}, for which each unentered room is a search opportunity with uncertain yield and delayed reward. Withholding unseen contents, not the room granularity itself, is what turns room selection into search. It also determines what the student is asked to imitate: object-level actions can depend on targets the student has not seen, whereas room-level actions depend only on information the student could acquire by exploring. Multi-room search preserves the key choice between a visible target and a room with uncertain, delayed reward.

Equal target values and unit primitive-action costs isolate these questions from target priorities and cost weights, while collection and uncertain exploration still compete for finite budget. Our focus is supervision and budget use within a fixed video VLN architecture: both Bellman experts supervise Uni-NaVid students that predict primitive actions from RGB history and language, and a greedy frontier expert and its supervised VLN provide coverage-oriented comparisons. Our goal is not to replace explicit-map exploration methods such as frontier planning; we use them as references for what explicit spatial memory achieves.
\else
This motivates comparing demonstrations that route among known targets with demonstrations that search regions for undiscovered targets. The empirical question is how the supervision schemes transfer to the same VLN.

Equal target values and unit primitive-action costs isolate these questions from target priorities and cost weights, while collection and uncertain exploration still compete for finite budget. Our focus is supervision and budget use within a fixed video VLN architecture. Multi-room search preserves the key choice between a visible target and a room with uncertain, delayed reward. In our room-level abstraction, doorway evidence estimates room category, and cross-scene priors predict discoveries while unseen object contents remain hidden. This representation specifies both the scale of planning and the information available for choosing where to search.

Using the same finite-budget Bellman principle, an \emph{object-level expert} routes among targets with the full map and locations, whereas a \emph{room-level expert} selects regions from observed connectivity, doorway evidence, and cross-scene priors. Their Uni-NaVid students predict primitive actions from RGB history and language; a greedy frontier expert and its supervised VLN provide coverage-oriented comparisons.
\fi

All three expert-supervised VLNs use matched scenes, budgets, and record counts; the two Bellman students also use identical optimization. Room-level supervision achieves the highest mean collection at every budget in ten held-out HM3D scenes despite lower expert return. Budget-blind controls distinguish countdown gains from supervision-source differences; countdown interventions test behavioral sensitivity. These findings motivate recording demonstration budgets and considering expert search behavior when choosing supervision.

\medskip
\noindent\textbf{Our contributions are:}
\begin{itemize}

\item We formulate \emph{budget-conditioned object search}, using distinct target collection to measure effective budget use and controlled countdown interventions to assess behavioral sensitivity to the budget.

\ifarxiv
\item We formalize expert design as a choice of abstraction $\varphi$ of the full environment. One budget-indexed Bellman recursion then spans fully informed routing and search under uncertainty, and the abstraction makes explicit which information enters the expert's actions and must be imitated by a partially observed student.

\item We instantiate a room-level Bellman expert that withholds the contents of unexplored rooms and estimates this recursion from strictly cross-scene transitions, providing budget-conditioned search demonstrations for a VLN.
\else
\item We instantiate a room-level Bellman expert that plans with uncertain target yields using observed evidence and cross-scene priors, providing budget-conditioned demonstrations for a partially observed VLN.
\fi

\item We compare room-level, fully informed object-level, and greedy-frontier supervision for Uni-NaVid on matched scenes, budgets, and record counts. Room-level supervision yields the highest mean collection, while higher expert return need not yield a stronger student.

\end{itemize}

\section{Related Work}

\subsection{Privileged and Planning Supervision}

\ifarxiv
A high-performing expert need not provide optimal supervision to a student
with partial information. ADVISOR~\cite{NEURIPS2021_9fc66491},
A2D~\cite{pmlr-v139-warrington21a}, and COSIL~\cite{pmlr-v205-nguyen23a}
reduce reliance on privileged actions or combine imitation with learning under
partial observability; recent work studies student realizability and
teacher--student alignment~\cite{kim2025distillingrealizablestudentsunrealizable,ICLR2026_398753dd}.
ADVISOR and A2D characterize the resulting imitation gap through the expert's
actions marginalized over information the student cannot observe, the
quantity we use in Sec.~\ref{sec:mismatch}.
Privileged planning also supervises navigation in Learning by
Cheating~\cite{learningbycheating}, shortest-path imitation in
SPOC~\cite{spoc}, and large-scale exploration~\cite{10476684}. These approaches
modify the imitation objective or adapt the expert to the student. We instead
vary the information available to the planning expert itself through its
scene abstraction, and study how the resulting supervision transfers to a
fixed video VLN, comparing fully informed object-level routing with room-level
search under uncertainty.
The two Bellman-supervised students share architecture and training settings.
Greedy-frontier supervision provides an additional comparison with
coverage-oriented exploration demonstrations.
\else
A high-performing expert need not provide optimal supervision to a student
with partial information. ADVISOR~\cite{NEURIPS2021_9fc66491}
and A2D~\cite{pmlr-v139-warrington21a}
reduce reliance on privileged actions or combine imitation with learning under
partial observability; recent work studies student realizability and
teacher--student alignment~\cite{kim2025distillingrealizablestudentsunrealizable,ICLR2026_398753dd}.
Privileged planning also supervises navigation in Learning by
Cheating~\cite{learningbycheating}, shortest-path imitation in
SPOC~\cite{spoc}, and large-scale exploration~\cite{10476684}. We study how
planning-based supervision transfers to a fixed video VLN, comparing fully
informed object-level routing with room-level search under uncertainty.
The two Bellman-supervised students share architecture and training settings.
Greedy-frontier supervision provides an additional comparison with
coverage-oriented exploration demonstrations.
\fi

\subsection{Vision-Language Navigation and Object Search}

Large-scale imitation and reinforcement learning support policies such as
ViNT and LM-Nav~\cite{vint,lmnav}, SPOC~\cite{spoc}, and video VLNs
NaVid~\cite{navid}, Uni-NaVid~\cite{uninavid}, and NaVILA~\cite{navila}.
Object search exploits geometry and semantics~\cite{sgnav,cognav,wmnav,muvla,vlfm},
\ifarxiv
with POMDPs modeling uncertainty over object locations~\cite{oopomdp,multires3dmos}.
Exploration uses rooms and doorways~\cite{Kim-2023-139130}, incremental
frontiers and hierarchical planning~\cite{zhou2021fuel}, or information gain
under mapping uncertainty~\cite{Suresh20icra}; cooperative systems further
combine exploration, mapping, and inspection in unknown
environments~\cite{xu2024costeffectivecooperativeexplorationinspection}.
\else
with POMDPs modeling location uncertainty~\cite{oopomdp}.
Exploration uses rooms and doorways~\cite{Kim-2023-139130} and hierarchical
frontier planning~\cite{zhou2021fuel}.
\fi
MultiON~\cite{multion}, GOAT-Bench~\cite{goatbench}, and long-horizon
VLN~\cite{longhorizonvln} extend task horizons. Whereas object search commonly
tests single-target success under a fixed limit~\cite{objectnavrevisited}, we
measure distinct discoveries and search-strategy changes across budgets.
Our comparisons examine supervision for a video VLN; explicit planning
methods supply demonstrations and reference trajectories.

\subsection{Budgeted and Budget-Conditioned Exploration}

\ifarxiv
Finite-resource decision making is classical in
orienteering~\cite{vansteenwegen2011orienteering}, constrained
MDPs~\cite{cmdp}, and informative path planning, where trajectories trade
information gain against limited time, distance, energy, or other
resources~\cite{activenerf,11128798,rig,popovic2020informative,pmlr-v205-cao23b,zhu2021onlineinformativepathplanning,moon2025iatigris}.
Our object-level Bellman expert solves an orienteering-style routing problem
over known targets; the room-level expert instead plans over expected yields
of unexplored regions. Recent planners use predicted maps and
pathwise information gain to reason about unseen regions along candidate
trajectories~\cite{baek2025pipe}. Resource dependence has also been
studied directly in learned exploration policies~\cite{chung2015learning},
alongside adaptive and frontier-based
exploration~\cite{adaptivesubmodularity,yamauchi1997frontier,chaplot2020ans}.

Decision Transformer~\cite{decisiontransformer} and reward-conditioned or
constrained policies~\cite{rcp,budgetedrl} condition on returns, horizons, or
constraints. Time Limits in RL~\cite{pardo2018timelimits} shows that omitting
the remaining horizon can alias states requiring different optimal actions;
navigation methods likewise incorporate budget, return, or termination
signals~\cite{star,batnav,ratenav,moon,controltransformer}. We focus on
\emph{exploration}: which regions to search and in what order as the budget
changes, and which expert supervision transfers this behavior to a video VLN.
\else
Finite-resource planning includes orienteering~\cite{vansteenwegen2011orienteering},
constrained MDPs~\cite{cmdp}, and informative paths trading information gain
against time, distance, or energy~\cite{activenerf,11128798,rig,pmlr-v205-cao23b,moon2025iatigris}.
Predicted maps and pathwise information gain support reasoning about unseen
regions~\cite{baek2025pipe}, alongside frontier-based
methods~\cite{yamauchi1997frontier,chaplot2020ans}.

Decision Transformer~\cite{decisiontransformer} and reward-conditioned or
constrained policies~\cite{rcp,budgetedrl} condition on returns, horizons, or
constraints. Time Limits in RL~\cite{pardo2018timelimits} explains horizon-induced
state aliasing; navigation methods incorporate budget, return, or termination
signals~\cite{batnav,ratenav,moon,controltransformer}. We focus on
\emph{exploration}: which regions to search and in what order as the budget
changes, and which expert supervision transfers this behavior to a video VLN.
\fi

\section{Budget-Conditioned Object Search}
\label{sec:task}

\subsection{Problem formulation}

An episode $e=(\mathcal{M},q_0,y,B)$ specifies a multi-room environment
$\mathcal{M}$---its complete geometry, semantics, and object instances---together
with the initial agent state, target category, and action budget.
Rooms organize exploration; the objective counts object instances and the
policy selects primitive actions. The budget can also represent time,
distance, energy, or another resource.

At step $t$, the policy $\pi(a_t\mid h_t,y,b_t)$ selects $a_t\in\mathcal{A}$
from observation--action history $h_t=(o_0,a_0,\ldots,o_t)$ and remaining
budget $b_t=B-\sum_{k=0}^{t-1}c(a_k)$, with positive action cost $c(a_t)$.
Here $\pi$ denotes any policy using these inputs; our learned instances are
Uni-NaVid students trained on expert actions (Sec.~\ref{sec:vln_results}).
Observations reveal $\mathcal{M}$ only partially, so different environments
may be indistinguishable under the same history.
\ifarxiv
$\mathcal{M}$ is therefore privileged: the simulator uses it to credit
discoveries, experts may access it only through the abstraction their design
permits (Sec.~\ref{sec:bellman_expert}), and the learned policy never
observes it.
\fi

Let $\mathcal{I}_y(\mathcal{M})$ contain the category-$y$ instances and
$\mathcal{F}_t\subseteq\mathcal{I}_y(\mathcal{M})$ those found by time $t$
under a policy-independent criterion. An instance is collected when it is
first credited by this criterion (Sec.~\ref{sec:experimental_setup}). The reward
$r_t=|\mathcal{F}_{t+1}\setminus\mathcal{F}_t|$ counts only newly credited
instances after action $a_t$; revisiting a credited instance adds no reward.
The objective is
\begin{equation}
\pi_B^\star
\in
\arg\max_{\pi}
\;
\mathbb{E}_{\pi}
\left[
\sum_{t=0}^{\tau-1} r_t
\right]
\qquad
\mathrm{s.t.}
\qquad
\sum_{t=0}^{\tau-1} c(a_t) \le B,
\label{eq:budget_objective}
\end{equation}
where $\tau$ denotes the termination time.

\ifarxiv
Changing $B$ changes the decision problem rather than merely the point at
which a fixed trajectory is truncated. Under a tight budget, actions with
high immediate expected reward may be preferred; with more budget, the agent
can invest in exploration whose reward is realized only later. The optimal
action therefore depends jointly on what has been observed and how much
budget remains, and omitting $b_t$ from the policy input aliases states that
require different actions~\cite{pardo2018timelimits}.
\else
Tight budgets may favor immediate reward; longer budgets can justify
exploration with delayed reward. Optimal actions therefore depend on both
observations and remaining budget, not merely a trajectory's truncation point.
\fi

Heterogeneous target values replace the count with
$r_t=\sum_{i\in\mathcal{F}_{t+1}\setminus\mathcal{F}_t}w_i$, $w_i\ge0$;
Eq.~\eqref{eq:budget_objective} already permits nonuniform positive costs.
We use $w_i=c(a_t)=1$. Extensions require corresponding expert reward/cost
estimates and policy inputs conveying the task specification. They preserve
our research questions, but persistence of the supervision advantage requires
further evaluation.

\subsection{Budget-conditioned evaluation}

\ifarxiv
Standard single-object search provides a sparse success signal: once one
instance is found, the episode counts as successful regardless of how
effectively the agent explored. Counting distinct discoveries as a function
of the available budget instead distinguishes policies beyond single-target
success and provides a denser evaluation of budget use.
\else
Counting additional discoveries distinguishes policies beyond single-target
success and provides a denser evaluation of budget use.
\fi

For an episode $e$, let $\mathcal{R}_y(e)$ denote the set of target instances
considered valid and reachable under the benchmark's embodiment and discovery
criterion. For a fixed evaluated policy, $\mathcal{F}_e(B)$ is the set of
distinct instances credited by termination of the rollout with initial budget
$B$, i.e., $\mathcal{F}_t$ at that rollout's termination time. Changing $B$
can change both the trajectory and its termination time, hence the discovered
set. We define normalized collection, the fraction of valid reachable
instances found, as
\begin{equation}
N_e(B)
=
\frac{
|\mathcal{F}_e(B) \cap \mathcal{R}_y(e)|
}{
|\mathcal{R}_y(e)|
},
\end{equation}
and report its average over the set $\mathcal{E}_B$ of matched evaluation
episodes run with budget $B$,
\begin{equation}
N(B)
=
\frac{1}{|\mathcal{E}_B|}
\sum_{e\in\mathcal{E}_B} N_e(B).
\label{eq:mean_score}
\end{equation}

Higher collection with more budget does not establish use of the budget
input: a fixed strategy may find more objects simply by running longer.
We therefore hold scene, target, start, model, and decoding fixed, vary only
the countdown, and examine trajectory divergence within the shared horizon.
These interventions test behavioral sensitivity; $N(B)$ measures performance.

\begin{figure*}[!t]
\centering
\includegraphics[width=\textwidth]{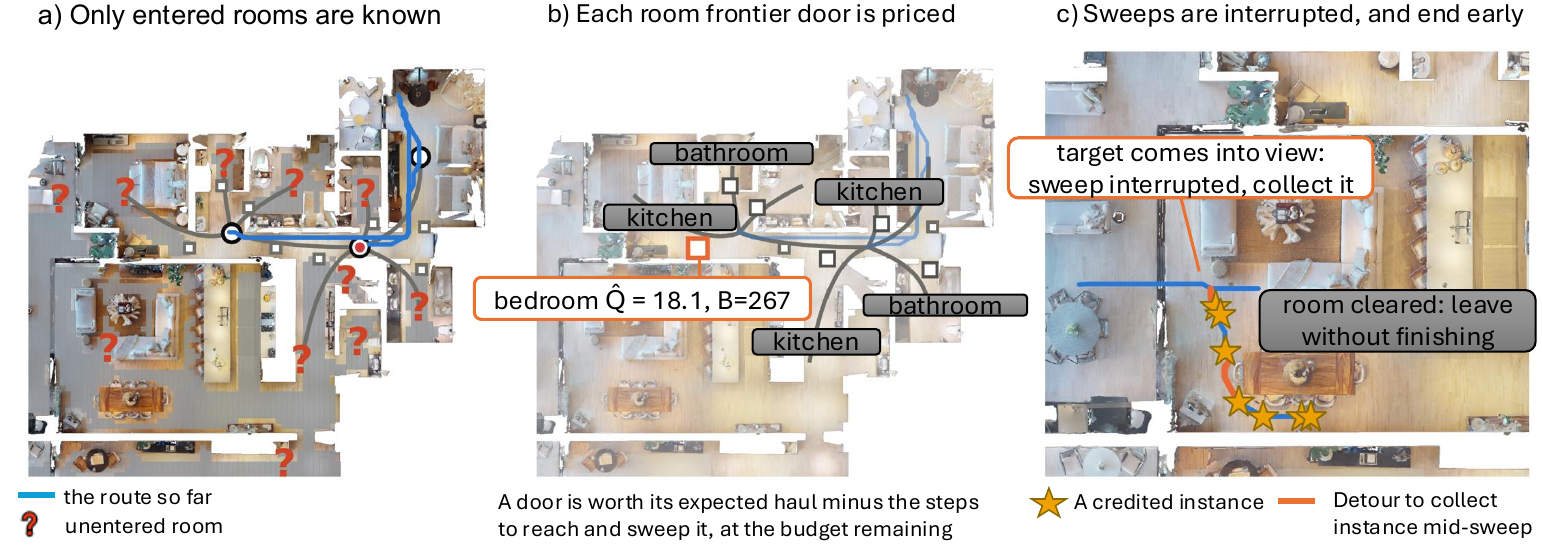}
\caption{\textbf{Room-level Bellman expert.}
(a) ``Known'' rooms have been entered; their recorded contents include only
discovered objects.
(b) Frontier rooms are valued using doorway-visible semantic evidence
and cross-scene Bellman estimates.
(c) Execution remains closed-loop: visible targets can interrupt a
sweep, after which the expert replans.}
\label{fig:mech}
\end{figure*}

\ifarxiv
\section{Abstract-State Bellman Experts}
\label{sec:bellman_expert}

A strong search expert can derive its advantage in two different ways: by
reasoning effectively from information that the learned policy can also
acquire, or by exploiting information that the learned policy will never
observe. The distinction matters for budget-conditioned search, where
demonstrations should teach not only where to go but how to trade immediate
collection against exploration whose payoff is uncertain. If an expert
observes the complete environment, including unseen target locations, object
search collapses to routing: the expert knows where the rewards are and only
needs to reach them efficiently. Such trajectories can be near-optimal, yet
they never demonstrate a decision made without knowing what lies behind a
door. We therefore treat the expert's information as an explicit design
choice. Both of our Bellman experts optimize the same budgeted objective; they
differ in the abstraction of the environment they plan over and in how much
of it is revealed to them.

\subsection{Bellman Planning over Scene Abstractions}
\label{sec:abstraction}

Recall that $\mathcal{M}$ denotes the complete, privileged environment
(Sec.~\ref{sec:task}). An expert does not plan over $\mathcal{M}$ directly,
but over an abstraction
\begin{equation}
z=\varphi(\mathcal{M}),
\label{eq:abstraction}
\end{equation}
where $\varphi$ is a design choice specifying which task-relevant structure of
$\mathcal{M}$ is retained, such as the locations of all target instances or a
room graph with room types and contents, and which details are discarded.
Like $\mathcal{M}$, $z$ is a property of the environment rather than of what
has been observed, and is therefore privileged. At time $t$, the expert's
information state $s_t$ summarizes the part of $z$ revealed to it so far,
together with the robot state and the set $\mathcal{F}_t$ of credited
instances. Two choices thus define an expert's information: what $\varphi$
retains, and what $s_t$ reveals.

Let $p(\mathcal{M})$ denote the distribution of environments and
$p(\mathcal{M}\mid s)$ the distribution over environments consistent with
information state $s$. Because multiple environments can remain consistent
with $s_t$, the expert evaluates actions through rewards and transitions
averaged under this distribution,
\begin{align}
R_\varphi(s,a)
&=
\mathbb{E}_{\mathcal{M}\mid s}
\big[\,r_t \mid \mathcal{M},s_t=s,a_t=a\,\big],
\label{eq:abstract_reward}\\
P_\varphi(s'\mid s,a)
&=
\mathbb{E}_{\mathcal{M}\mid s}
\big[\Pr(s_{t+1}=s'\mid \mathcal{M},s,a)\big],
\label{eq:abstract_transition}
\end{align}
where $\mathbb{E}_{\mathcal{M}\mid s}$ denotes expectation over
$\mathcal{M}\sim p(\mathcal{M}\mid s)$. Here $a$ may be a temporally extended action, such as visiting a target or
entering and searching a room, with $c(a)$ its total primitive-action cost and
$r_t$ the reward accumulated over its primitive steps. For remaining budget
$b$, the corresponding Bellman recursion is
\begin{align}
Q_\varphi^\star(s,b,a)
&=
R_\varphi(s,a)
+
\mathbb{E}_{s'\sim P_\varphi(\cdot\mid s,a)}
\left[
V_\varphi^\star(s',b-c(a))
\right],
\label{eq:abstract_bellman_q}
\\
V_\varphi^\star(s,b)
&=
\max_{a:c(a)\le b} Q_\varphi^\star(s,b,a),
\qquad
V_\varphi^\star(s,0)=0.
\label{eq:abstract_bellman_v}
\end{align}
Both of our Bellman experts implement
Eqs.~\eqref{eq:abstract_bellman_q}--\eqref{eq:abstract_bellman_v}. They
differ in $\varphi$, in what $s_t$ reveals, and therefore in the distribution
$p(\mathcal{M}\mid s_t)$ that the expectations average over.

\subsection{A Spectrum of Abstractions}
\label{sec:spectrum}

The choice of $\varphi$ and of what $s_t$ reveals determines the uncertainty
the expert must reason over, and spans a spectrum of information access.
Table~\ref{tab:abstractions} contrasts the two Bellman experts with the
student they supervise.

\noindent\textbf{Revealed targets: object-level routing.}
If $\varphi$ retains the map and every target instance, and $z$ is revealed
in full at $t=0$, then $s_t$ determines all future rewards. The expectations
in Eqs.~\eqref{eq:abstract_reward}--\eqref{eq:abstract_transition} collapse,
and the recursion can be written as deterministic budgeted routing over known
targets, an orienteering problem~\cite{vansteenwegen2011orienteering}:
\begin{equation}
V_{\mathrm{obj}}(s,b)
=
\max\Big(0,
\max_{i\in\mathcal{K}_b(s)}
\big[1+V_{\mathrm{obj}}(s\oplus i,b-\kappa_i)\big]\Big),
\label{eq:object_bellman}
\end{equation}
where $\kappa_i=\kappa(s,i)$ is the primitive-action cost of reaching a
crediting pose of instance $i$ from $s$,
$\mathcal{K}_b(s)=\{i\in\mathcal{I}_y(\mathcal{M})\setminus\mathcal{F}(s):
\kappa_i\le b\}$ contains the uncredited instances reachable within budget
$b$, $\mathcal{F}(s)$ is the set of credited instances, and $s\oplus i$ is
the state after crediting $i$. Decisions concern which known
instance to visit next; ``fully informed'' refers to knowledge of every target
location, including those outside the camera's view. This is our
object-level Bellman expert: it has nothing left to search for.

\noindent\textbf{Observed geometry only: coverage exploration.}
At the other extreme, if $\varphi$ retains only geometry and the expert has no
informative prior on what unexplored regions contain, it has little basis for
distinguishing them and plans essentially for coverage, as in frontier and
receding-horizon exploration~\cite{yamauchi1997frontier,bircher2016receding}.
Such planners act only on information a student can also acquire, but give
weak guidance about where distant rewards are likely to be.

\noindent\textbf{Intermediate abstractions: search with priors.}
Between these extremes, $\varphi$ can retain coarse structure of unseen
regions, namely their connectivity and likely semantics, while $s_t$ withholds
their realized contents. Actions are then valued by expected return over
environments consistent with the current information, with
$p(\mathcal{M}\mid s_t)$ supplying cross-scene regularities such as which room
types tend to contain the target category. An action with little immediate
reward may still be optimal if it reaches a promising region or resolves
uncertainty that enables future reward, and whether it is worthwhile depends
on the remaining budget $b$. Exploration can thus be viewed as acting to
resolve the task-relevant uncertainty left by the chosen abstraction. Our
room-level Bellman expert
(Secs.~\ref{sec:room_instantiation}--\ref{sec:closed_loop}) occupies this
regime.

\noindent\textbf{Granularity versus information access.}
These are distinct axes. Planning over rooms does not by itself create
uncertainty: a room-level expert given the realized contents of every room
would face a deterministic problem like Eq.~\eqref{eq:object_bellman}, with
rooms rather than instances as decision units. It is withholding unrevealed
contents that turns room selection into search. Our two Bellman experts differ
along both axes, which our experiments do not separate
(Sec.~\ref{sec:discussion}).

\begin{table}[t]
\centering
\caption{Information available to each Bellman expert and to the student
they supervise. $\mathcal{M}$ is the full environment,
$z=\varphi(\mathcal{M})$ the expert's abstraction, and $s_t$ the part of $z$
revealed at time $t$.}
\label{tab:abstractions}
\footnotesize
\setlength{\tabcolsep}{2.5pt}
\renewcommand{\arraystretch}{1.15}
\begin{tabular}{@{}>{\raggedright\arraybackslash}p{0.18\columnwidth}
>{\raggedright\arraybackslash}p{0.24\columnwidth}
>{\raggedright\arraybackslash}p{0.27\columnwidth}
>{\raggedright\arraybackslash}p{0.23\columnwidth}@{}}
\hline
 & Object-level expert & Room-level expert & VLN student \\
\hline
Abstraction $z$
  & map and all target instances
  & room graph with room types and contents
  & none; raw RGB \\
Revealed in $s_t$
  & all of $z$, from $t=0$
  & observed rooms and connectivity, discovered objects, doorway evidence
  & RGB history $h_t$ and countdown $b_t$ \\
Withheld
  & ---
  & contents of unentered rooms
  & everything unseen \\
Planning
  & deterministic routing, Eq.~\eqref{eq:object_bellman}
  & expected return from cross-scene transitions, Eq.~\eqref{eq:bellman_q}
  & none; imitates expert actions \\
Decision unit
  & next target instance
  & next frontier room
  & next four primitives \\
Actions depend on unseen targets
  & yes & no & no \\
\hline
\end{tabular}
\end{table}

\subsection{Abstraction and Expert--Student Information Mismatch}
\label{sec:mismatch}

Each expert $\pi_E(a\mid s_t,b_t)$ supplies action labels for a separate
Uni-NaVid student $\pi_\theta(a\mid h_t,y,b_t)$ trained with the same
imitation objective. At inference, students act from RGB history and
language, including the countdown, without querying an expert or running its
planner. With a cross-entropy imitation loss on expert-visited states, the
best student in an unrestricted policy class matches the expert's actions
marginalized over the expert information that its own history does not
determine~\cite{NEURIPS2021_9fc66491,pmlr-v139-warrington21a}:
\begin{equation}
\pi_\theta^\star(a\mid h_t,y,b_t)
=
\mathbb{E}_{s_t\sim p(\cdot\mid h_t,y,b_t)}
\big[\pi_E(a\mid s_t,b_t)\big].
\label{eq:implicit_policy}
\end{equation}
When $s_t$ is determined by $h_t$, the student can reproduce the expert.
Otherwise it learns a mixture of the expert's actions across environments it
cannot distinguish, and for budget-conditioned search such a mixture need not
be a coherent search strategy. For example, from an identical start view the
object-level expert may turn left in one scene and right in another depending
on where unseen targets lie; the student receives identical inputs and can
only learn their average.

The abstraction controls which information enters this marginalization. The
object-level expert's $s_t$ contains every target location from $t=0$, so
Eq.~\eqref{eq:implicit_policy} averages routes toward targets the student has
not seen. The room-level expert withholds unrevealed contents from its own
decisions, so its actions depend only on information a student could in
principle acquire by exploring or learn as a prior from training scenes. A
residual mismatch remains: the room-level expert maintains an explicit room
graph and reads observed room types and objects from simulator annotations,
which the student must infer from RGB history. Rather than adapting the
learning objective or the expert to a particular student, we use $\varphi$ to
control the expert's information, seeking an abstraction that retains enough
structure for long-horizon planning while requiring the expert to reason over
uncertainty that the student must also resolve through exploration.

This argument motivates the room-level design but does not predict the
outcome: a better-aligned expert may also be a weaker one, as the object-level
expert's higher return shows (Sec.~\ref{sec:expert_analysis}), and the two
experts differ in more than information access. Section~\ref{sec:vln_results}
tests the resulting supervision empirically.
\else
\section{Room-Level Bellman Expert}
\label{sec:bellman_expert}
To generate search demonstrations, we instantiate an expert that retains spatial and
semantic structure for planning while discovering unseen targets through
budget-dependent exploration.

The abstraction $z=\varphi(\mathcal{M})$ retains task-relevant structure of
the full environment, e.g., target locations or a room graph; like
$\mathcal{M}$, $z$ is privileged. The expert's information state $s_t$
summarizes what has been revealed about $z$ to the expert. Multiple
environments can remain compatible with $s_t$, giving expected rewards and
transitions
\begin{align}
R_\varphi(s,a)
&=
\mathbb{E}[r_t \mid s_t=s,a_t=a], \\
P_\varphi(s' \mid s,a)
&=
\Pr(s_{t+1}=s' \mid s_t=s,a_t=a).
\end{align}
For remaining budget $b$, the corresponding Bellman recursion is
\begin{align}
Q_\varphi^\star(s,b,a)
&=
R_\varphi(s,a)
+
\mathbb{E}_{s'\sim P_\varphi(\cdot\mid s,a)}
\left[
V_\varphi^\star(s',b-c(a))
\right],
\label{eq:abstract_bellman_q}
\\
V_\varphi^\star(s,b)
&=
\max_{a:c(a)\le b} Q_\varphi^\star(s,b,a),
\qquad
V_\varphi^\star(s,0)=0.
\label{eq:abstract_bellman_v}
\end{align}

\noindent\textbf{Shared planning principle across scene abstractions.} Both Bellman experts optimize immediate collection plus continuation value under a finite budget. The object-level expert routes among known targets using the full map and true locations. The room-level expert represents unexplored rooms through observed connectivity and doorway-based category estimates. Cross-scene priors predict discoveries, execution costs, and successor states while unseen objects and locations remain unknown. This abstraction specifies both planning units and information access. The room-level implementation estimates Bellman values from cross-scene transitions, as detailed below; it does not claim an exact solution of the underlying partially observable search problem.

Each expert supplies action labels to train a separate Uni-NaVid student with the same imitation objective. At inference, students act from RGB history and language, including the countdown, without querying an expert or running its planner. The room-level expert additionally maintains an explicit room graph and semantic state; expert and student inputs therefore differ.
\fi

\subsection{Room-Level Instantiation}
\label{sec:room_instantiation}

Room entry incurs travel and search costs, reveals targets, and may expose
further regions; semantics provide target-yield priors. This captures the
collection--exploration tradeoff without requiring a room representation in
the task objective or student's action space. Figure~\ref{fig:mech} shows
entered rooms and adjacent unentered candidates, evaluated by expected
yield and the budget needed to reach and search them.

Our room-level $\varphi$ maps $\mathcal{M}$ to a room graph with room types
and contents. The information state $s_t$ reveals only its observed part,
withholding the geometry, contents, and target locations of unexplored rooms:
it contains the observed room graph
$G_t^{\mathrm{obs}}=(V_t^{\mathrm{obs}},E_t^{\mathrm{obs}})$, robot state,
observed room types and discovered contents, and frontier-doorway evidence.
Observed room types and objects come from simulator ground-truth annotations,
exposed as rooms are entered or objects become visible, rather than inferred
from RGB. Entering a room does not reveal its complete object inventory;
only visible objects become available for direct approach.

The frontier set $\mathcal{U}(s_t)$ contains adjacent unentered rooms on the
current floor with a doorway reachable along a finite geodesic. An unentered
room may have objects visible through its doorway; this limited evidence
does not reveal its complete contents. Such objects provide a room-type
estimate through the cross-scene prior $P(\rho\mid o)$ learned from
Matterport3D~\cite{mp3d}, where $\rho$ is room type and $o$ is doorway-visible
object evidence. The expert uses this estimate to predict discoveries in the
unobserved interior. An estimated full sweep exceeding the budget does not
exclude a frontier; the budget instead filters transition outcomes below.

\subsection{Cross-Scene Bellman Estimation}
\label{sec:cross_scene}

We estimate the unknown reward and transition models from a bank of
room-transition records from 181 HM3D scenes. Each record
$\xi_i=(s_i,v_i,r_i,d_i,s_i')$ stores the information state before selecting
frontier room $v_i$, the newly credited target counts $r_i$, the realized
primitive-action cost $d_i$, and the information state after execution.
Thus, $r_i^y$ is a room-transition reward for category $y$, accumulating the
primitive-step rewards defined in Sec.~\ref{sec:task}. These outcomes supply
empirical samples for Eqs.~\eqref{eq:abstract_bellman_q}--\eqref{eq:abstract_bellman_v};
the doorway prior supplies the candidate's estimated room type, not its
realized reward. This bank supports expert planning; the video/action records
used to train the VLN are described in Sec.~\ref{sec:vln_results}.

For a query $(s,v)$, we retrieve transitions with matching abstract
context--action signatures and use weighted nearest neighbors when exact
matches are sparse. Retrieval is strictly cross-scene. The
source scenes are split into five folds, and planning in a scene excludes
every transition from that scene's entire fold. All benchmark scenes are excluded from the evaluation transition
bank.
\ifarxiv
Each retrieved transition is thus one outcome of the same abstract action in
another environment consistent with the same information state, i.e., an
empirical sample from $p(\mathcal{M}\mid s)$ in
Eqs.~\eqref{eq:abstract_reward}--\eqref{eq:abstract_transition}. Strict
cross-scene retrieval ensures that these samples never come from the realized
contents of the scene being planned: the prior is learned from other
environments rather than read off the current one.
\fi

Retrieved transitions jointly sample reward, execution cost, and successor
context. Let $\mathcal{N}_b(s,v)$ contain those with $d_i\le b$, with
weights $\alpha_i\ge0$ normalized to $\sum_{i\in\mathcal{N}_b(s,v)}\alpha_i=1$.
For category $y$, the empirical backup combines immediate reward and
continuation return:
\begin{equation}
\widehat{Q}_y(s,b,v)
=
\sum_{i\in\mathcal{N}_b(s,v)}
\alpha_i
\left[
r_i^y
+
\widehat{V}_y(s_i',b-d_i)
\right],
\label{eq:bellman_q}
\end{equation}
where $r_i^y$ is the immediate room-transition reward for category $y$ and
$\widehat{V}_y(s_i',b-d_i)$ estimates the best subsequent return after
spending $d_i$ steps. The state value selects the highest estimated return
over frontier rooms and a zero-valued stopping option:
\begin{equation}
\widehat{V}_y(s,b)
=
\max
\left(
0,\;
\max_{v\in\mathcal{U}(s)}
\widehat{Q}_y(s,b,v)
\right),
\qquad
\widehat{V}_y(s,0)=0.
\label{eq:bellman_v}
\end{equation}

The hats denote estimates computed recursively under the empirical model.
Filtering by $d_i\le b$ and renormalizing conditions on outcomes that fit
the budget; it does not model the probability or reward of exceeding it.

A room with no admissible transition has undefined $\widehat{Q}_y$ and is
excluded from the maximization, but remains eligible for the fallback below.

Continuation enables non-myopic exploration: a low-reward connector room
may expose promising regions, yet be unhelpful under a tight budget.
Positive transition costs permit evaluation in increasing order of $b$.

Doorway evidence thus estimates room type; retrieved outcomes turn this
semantic expectation into reward, cost, and continuation estimates. A room's
ranking depends on budget-dependent returns, not its semantic promise alone.

\subsection{Closed-Loop Execution}
\label{sec:closed_loop}

Primitive execution is closed-loop: a newly visible uncredited target
within 5\,m and line of sight interrupts the room sweep for approach-and-collect.
The executor uses observed targets and coverage to determine when its room
search is complete; it does not consult a ground-truth room inventory or
target count.
Figure~\ref{fig:mech}(c) shows a mid-sweep detour and departure with most
sweep viewpoints unvisited. Alternating scan directions at successive
viewpoints reduces the left:right turning ratio from 2:1 to 1.2:1.

If no frontier has positive estimated value or admissible evidence, the
expert enters the nearest unexplored frontier as a curiosity fallback.
The zero branch in Eq.~\eqref{eq:bellman_v} is a value-estimation option,
not an execution rule requiring termination when positive evidence is absent.
After every room transition or interruption it updates $s_t$ and
replans. If all frontiers are exhausted while budget remains, it
re-sweeps visited rooms, providing demonstrations at horizons longer
than required to explore the scene.

Unexplored contents and target locations are used only after execution to
construct transitions, never for action selection. The abstraction merges
distinct histories and the model uses finite cross-scene data, so this is an
\emph{empirical room-level Bellman expert}, without a Bayes-optimality claim
for the underlying partially observable search problem.
\ifarxiv
In the notation of Sec.~\ref{sec:abstraction}, the expert conditions on the
abstract state $s_t$ rather than on its full history, so
$p(\mathcal{M}\mid s_t)$ can differ from the posterior given that history,
and the expectations in
Eqs.~\eqref{eq:abstract_reward}--\eqref{eq:abstract_transition} are
approximated from finite cross-scene samples.
\fi

\begin{figure}[!t]
\centering
\includegraphics[width=\linewidth]{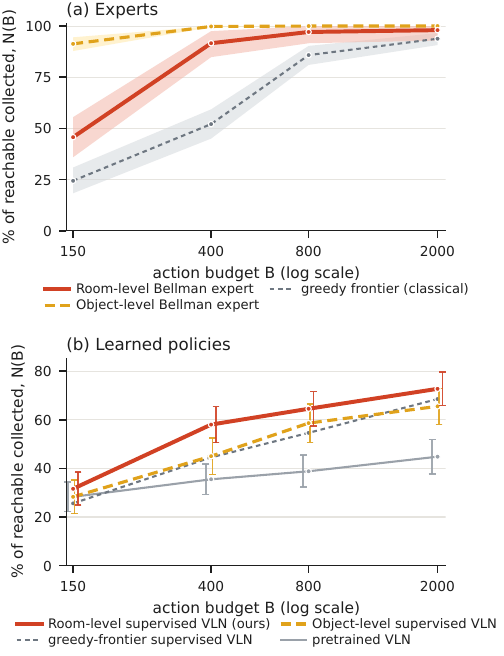}
\caption{\textbf{Expert return and supervision transfer on the frozen benchmark.}
\textbf{(a)} Expert collection $N(B)$ with 95\% bootstrap intervals:
object-level Bellman leads room-level Bellman, which leads unprivileged
frontier baselines. \textbf{(b)} Mean collection after matched Uni-NaVid
fine-tuning reverses the ordering of the two Bellman experts. Greedy frontier is a
non-learned reference; Table~\ref{tab:vln_results} additionally reports
greedy-frontier-supervised VLN performance. Budget-dependent margins and
long-horizon revisiting are discussed in Sec.~\ref{sec:vln_results}.}
\label{fig:curves}
\end{figure}

\section{Experiments}
\label{sec:experiments}

We test budget-dependent VLN behavior through collection scores and
countdown interventions, and supervision transfer through matched training
comparisons. Experts and learned policies share the objective, budgets, and
discovery criterion. The two Bellman experts also share the budgeted
optimization principle, applied to known targets or expected room yields
according to their available information (Sec.~\ref{sec:bellman_expert});
greedy frontier provides a coverage-oriented baseline.
We examine demonstrations before learned policies, using the held-out HM3D
protocol below.

\subsection{HM3D Experimental Setup}
\label{sec:experimental_setup}

\noindent\textbf{Environment and action space.}
We instantiate budget-conditioned object search in multi-room
HM3D~\cite{hm3d} indoor scenes in Habitat~\cite{habitat} with
$\mathcal{A}={\textsc{Forward},\textsc{Left},\textsc{Right},\textsc{Stop}}$,
where \textsc{Forward} moves $0.25$\,m and \textsc{Left}/\textsc{Right} rotate
$30^\circ$. Each primitive costs one budget unit, and episodes terminate on
\textsc{Stop} or budget exhaustion. We evaluate
$B\in{150,400,800,2000}$. Learned VLNs receive the remaining budget only
through language, e.g., ``Find as many chairs as possible. You have $b_t$
steps remaining,'' with no dedicated numerical budget input. Their
observations consist of RGB video history; the full map, room graph,
and ground-truth locations of unseen targets are not provided to them.

\smallskip
\noindent\textbf{Discovery criterion.}
We define an object as found using a policy-independent geometric referee.
A target instance is credited when the agent is within $1.5$\,m of its
bounding extent, within $1.5$\,m vertically, and its bearing is within
$39.5^\circ$ of the camera optical axis. This bound is half of the camera's
$79^\circ$ horizontal field of view. The criterion measures target coverage
within the camera's nearby viewing region. It is applied identically to every
policy and requires no separate recognition report or collection action. Each instance is
credited at most once. We exclude instances in the starting room and those
without any valid crediting pose geodesically reachable from the start.
Reachability is computed directly on the HM3D navmesh rather than the room
graph, allowing, e.g., an instance visible through a doorway to be credited
without entering its semantic room.

\smallskip
\noindent\textbf{Matched evaluation protocol.}
We evaluate all methods on 10 held-out HM3D-val scenes excluded from both VLN
fine-tuning and room-level Bellman transition data. The benchmark contains 50
admission-verified $(\mathrm{scene},\mathrm{category})$ pairs with two
deterministic starts each. Scene, category, initial position, heading, and
budget are matched across methods, giving 100 episodes per budget.
Table~\ref{tab:vln_results} reports mean normalized collection $N(B)$
(Sec.~\ref{sec:task}) with the half-width of a 95\% confidence interval (CI).
CIs are percentile bootstraps over evaluation episodes: for each method and
budget we resample the matched episode set 10{,}000 times with replacement and
take the 2.5th and 97.5th percentiles of the resampled mean. The episode is the
resampling unit, so scene, category, and start are resampled together.
Fig.~\ref{fig:curves}(a) draws these intervals as bands and
Fig.~\ref{fig:curves}(b) as error bars, offset horizontally for legibility.
We also examine cross-budget trajectory divergence, revisiting, and
monotonicity violations, where a larger budget yields fewer targets.

\subsection{Expert Supervision}
\label{sec:expert_analysis}

\noindent\textbf{Expert performance.}
The object-level Bellman expert solves budgeted routing with the full map
and target locations; the room-level expert follows the information
constraints in Sec.~\ref{sec:bellman_expert}. The greedy frontier expert
provides a coverage-oriented exploration baseline with explicit spatial
memory. We evaluate both its direct execution and the VLN trained on its
demonstrations under the same held-out protocol and action budgets.

The object-level expert leads at every budget, most strongly at tight
budgets (Fig.~\ref{fig:curves}(a)). Both Bellman experts act until the budget
is exhausted: after collecting all targets, the object-level expert re-sweeps
the environment, while the room-level expert re-sweeps visited rooms after
exhausting frontiers. These continuation behaviors are part of the
respective supervision schemes. Re-sweeping ensures every demonstration
reaches its assigned budget, controlling the distribution of remaining
budgets in training records across experts despite different completion times.

\ifarxiv
The gap between the two Bellman experts measures what revealing target
locations is worth to a planner: 45.6 points at $B=150$ (91.3\% versus
45.7\%), shrinking to 8.2, 2.9, and 2.0 points at $B=400/800/2000$
(Table~\ref{tab:vln_results}). Knowing where targets are matters most when
little budget remains to recover from an uninformed first choice. This
advantage is largest exactly where decisions depend most on target locations
that the student cannot observe (Sec.~\ref{sec:mismatch}), so expert return
alone does not determine how useful demonstrations are for learning.
\fi

\noindent\textbf{Budget-dependent demonstrations.}
\ifarxiv
High collection alone does not establish that the room-level expert changes
its strategy with the available budget.
\fi
Varying only $B$ at matched scenes, target categories, and starts, 55\% of
691 admitted training pairs select a different first room across budgets;
96\% are monotone in targets collected.

At equal executed-action counts, mean cumulative collection for $B=150$
leads larger-budget demonstrations for about 131 steps, by up to 7.0 points,
before saturating. This controls for additional execution time: tight budgets
favor immediate yield, while larger ones justify delayed-reward transfers. Thus, the demonstrations contain budget-dependent decisions rather than
truncations of a common route.

\subsection{Learning a Budget-Conditioned VLN}
\label{sec:vln_results}

\noindent\textbf{Fine-tuning setup.}
Bellman-expert demonstrations are converted to Uni-NaVid's native training format
using its fixed inference cadence: after every two executed primitives, the
model is queried for the next four actions. Each record contains the video
history up to time $t$, the expert's next four primitives, and a
natural-language countdown with $B-t$ steps remaining. The budget is provided
only through language; we introduce no numerical budget input or architectural
modification.

The room-level Bellman supervision has 388,199 records: 214,562 from 89 HM3D
and 173,637 from 59 Matterport3D scenes, at budgets 50, 150, 400, 800, 1200,
and 2000. A deterministic 20\% of records omit the countdown. We train directly on this
mixture without contrast mining or reweighting. For each Bellman supervision
scheme, we also train a \emph{budget-blind} control using the same video/action
records, with the countdown removed from every record. Architecture,
optimization, and training steps are fixed; only the training countdown text
differs. The controls retain budget-dependent trajectories but receive no
explicit countdown during training.

We otherwise follow the Uni-NaVid stage-2 recipe: the vision encoder is frozen,
the language model and projector are trained, the learning rate is $10^{-5}$,
and training lasts one epoch, 3{,}032 steps at effective batch size 128,
taking approximately 62 hours on four RTX PRO 6000 Blackwell GPUs.
The object-level, room-level, and greedy-frontier students use the same
scenes, budgets, and record count. The two Bellman students also share
architecture and optimization; only their expert supervision differs.

\noindent\textbf{Learned-policy performance.}
Table~\ref{tab:vln_results} compares experts and learned policies.
The non-learned greedy frontier uses explicit spatial memory, whereas VLNs
retain history implicitly in context. Its direct execution provides a
long-horizon coverage reference; its supervised VLN tests transfer of these
demonstrations to RGB history. The central comparison concerns supervision
within the Uni-NaVid architecture.

\begin{table}[t]
\centering
\caption{Budget-conditioned object-search performance (mean $\pm$ 95\% CI half-width):
demonstration experts and Uni-NaVid supervision variants.}
\label{tab:vln_results}
\small
\setlength{\tabcolsep}{3.5pt}
\resizebox{\columnwidth}{!}{%
\begin{tabular}{lcccc}
\hline
Policy & $B=150$ & $B=400$ & $B=800$ & $B=2000$ \\
\hline
\multicolumn{5}{l}{\textbf{(a) Expert policies}} \\
\hline
Greedy frontier expert
    & 24.4 {\scriptsize $\pm$6.3} & 52.1 {\scriptsize $\pm$7.1} & 85.8 {\scriptsize $\pm$4.6} & 93.8 {\scriptsize $\pm$2.9} \\
Object-level Bellman expert
    & 91.3 {\scriptsize $\pm$3.4} & 99.8 {\scriptsize $\pm$0.2} & 100.0 {\scriptsize $\pm$0.0} & 100.0 {\scriptsize $\pm$0.0} \\
Room-level Bellman expert (ours)
    & 45.7 {\scriptsize $\pm$9.8} & 91.6 {\scriptsize $\pm$6.4} & 97.1 {\scriptsize $\pm$4.2} & 98.0 {\scriptsize $\pm$3.5} \\
\hline
\multicolumn{5}{l}{\textbf{(b) Learned VLN policies}} \\
\hline
Pretrained VLN
    & 28.3 {\scriptsize $\pm$6.1} & 35.5 {\scriptsize $\pm$6.2} & 38.8 {\scriptsize $\pm$6.6} & 44.8 {\scriptsize $\pm$7.0} \\
VLN trained on greedy frontier expert
    & 25.6 {\scriptsize $\pm$6.0} & 44.5 {\scriptsize $\pm$7.3} & 54.6 {\scriptsize $\pm$7.5} & 68.5 {\scriptsize $\pm$7.3} \\
VLN trained on object-level expert
    & 28.3 {\scriptsize $\pm$6.2} & 45.0 {\scriptsize $\pm$7.3} & 58.6 {\scriptsize $\pm$7.5} & 65.5 {\scriptsize $\pm$7.4} \\
\quad budget-blind training
    & 25.5 {\scriptsize $\pm$6.0} & 42.2 {\scriptsize $\pm$7.2} & 55.0 {\scriptsize $\pm$7.5} & 61.8 {\scriptsize $\pm$7.5} \\
\textbf{VLN trained on room-level expert (ours)}
    & \textbf{31.6} {\scriptsize $\pm$6.8} & \textbf{58.0} {\scriptsize $\pm$7.4} & \textbf{64.5} {\scriptsize $\pm$7.1} & \textbf{72.7} {\scriptsize $\pm$6.8} \\
\quad budget-blind training
    & 28.5 {\scriptsize $\pm$6.2} & 51.6 {\scriptsize $\pm$7.5} & 59.5 {\scriptsize $\pm$7.5} & 63.5 {\scriptsize $\pm$7.5} \\
\hline
\end{tabular}%
}

\vspace{2pt}
\parbox{\columnwidth}{%
\footnotesize\raggedright
Panel (a): experts. Panel (b): pretrained Uni-NaVid or fine-tuned on the
named expert; indented rows repeat that supervision with the countdown
removed from every training record. Values are mean $N(B)$ (\%)
$\pm$ the half-width of the 95\% confidence interval. $N(B)$ is the percentage
of distinct valid, reachable targets found within $B$ primitives, averaged
across the 100 matched episodes. Higher is better;
bold marks the highest mean among evaluated VLNs at each budget.
}
\end{table}

At $B=150/400/800/2000$, room-level supervision improves over the pretrained
VLN by 3.3/22.5/25.7/27.9 points, over object-level supervision by
3.3/13.0/5.9/7.2, and over greedy-frontier supervision by 6.0/13.5/9.9/4.2;
its margin over the pretrained VLN widens with budget. At $B=2000$, it yields 72.7\% collection,
versus 68.5\% for greedy-frontier supervision, 65.5\% for object-level
supervision, and 44.8\% for the pretrained VLN. All VLNs beat the non-learned greedy
frontier expert at $B=150$, only room-level supervision does at $B=400$, and
the frontier expert leads every VLN from $B=800$ onward
(Table~\ref{tab:vln_results}).

\ifarxiv
One plausible explanation for this crossover is that learned semantic priors
help prioritize promising regions when only a few actions are available,
whereas a coverage-oriented frontier strategy may spend scarce budget on
regions regardless of target likelihood. At longer horizons, explicit spatial
memory dominates: greedy frontier tracks explored and unexplored regions
directly, while the VLN must retain this information implicitly in its
growing context. Consistent with this explanation, at $B=800$ and $B=2000$,
92--93\% of the room-level-supervised VLN's forward actions revisit previously
traversed cells. Short horizons thus appear to favor semantic prioritization,
and long horizons reliable coverage and memory.
\fi

The ordering of the two Bellman experts therefore reverses after distillation:
higher object-level expert return is associated with lower student collection, while room-level
performance exceeds greedy frontier before and after distillation. At
$B=2000$, the greedy-frontier-supervised VLN also beats the object-level
student (68.5\% versus 65.5\%), further showing that expert-return rankings
need not transfer; this comparison does not isolate Bellman planning from
other demonstration differences.

\ifarxiv
The comparison evaluates object-level routing and room-level search as
complete supervision schemes, including their abstractions and execution
behaviors. Information alignment is one possible explanation for the
reversal~\cite{NEURIPS2021_9fc66491,pmlr-v139-warrington21a}. By
Eq.~\eqref{eq:implicit_policy}, a student imitating the object-level expert
learns routes averaged over target placements it cannot see, whereas
room-level demonstrations search for unseen targets using semantic
expectations that depend only on evidence and priors the student could also
acquire. Under this view, the information that makes the object-level expert
stronger is the same information that makes its actions harder to imitate.
The empirical result establishes that expert-return rankings need not
transfer; it does not attribute the difference to information access alone,
because the experts also differ in planning granularity and execution
(Sec.~\ref{sec:discussion}).
\else
The comparison evaluates object-level routing and room-level search as
complete supervision schemes, including their abstractions and execution
behaviors. Information alignment is one possible explanation for the
reversal~\cite{NEURIPS2021_9fc66491,pmlr-v139-warrington21a}:
room-level demonstrations search for unseen targets using semantic
expectations, whereas object-level routes use their known locations. The
empirical result establishes that expert-return rankings need not transfer;
it does not attribute the difference to information access alone.
\fi

\noindent\textbf{Budget-blind training.}
Within each expert's fixed demonstrations, countdown annotations improve
mean collection at
$B=150/400/800/2000$ by 2.8/2.8/3.6/3.7 percentage points for object-level
and 3.1/6.4/5.0/9.2 for room-level supervision (Table~\ref{tab:vln_results}).
With countdowns omitted from both students' training, room-level supervision
still leads by 3.0/9.4/4.5/1.7 points. Thus, its advantage persists without
explicit budget annotations, while countdown gains depend on supervision
source. These comparisons distinguish supervision-source and countdown
effects while retaining budget-dependent expert trajectories. The next
intervention tests countdown use at inference.

\noindent\textbf{Controlled countdown intervention.}
To test use of the budget input independently of collection scores, matched
probes fix the initial observation, model, and deterministic decoding while
changing only the countdown sentence.

Changing $B=150$ to $B=2000$ changes the room-level Bellman-supervised VLN's
trajectory on all six held-out-scene pairs, with divergence after 2--11
actions. Collection increases on every pair:
$2\!\rightarrow\!8$,
$4\!\rightarrow\!7$,
$1\!\rightarrow\!10$,
$3\!\rightarrow\!12$,
$1\!\rightarrow\!7$, and
$4\!\rightarrow\!11$.
Same-budget repeats reproduce the action sequence. Divergence before either
budget expires establishes countdown sensitivity; collection gains also
reflect additional available actions and do not alone establish more
efficient allocation. In the mesh-rendered probe of Fig.~\ref{fig:teaser},
the $B=50/150/800$ routes diverge within the first few actions from the same
start and collect 1/1/16 of 16 chairs.

Across 206 mined contrast pairs, balanced first-action accuracy against the
room-level expert's budget-conditioned actions rises from 0.344 pretrained to
0.572 after room-level supervision. Yet on the 84 pairs where the expert
changes its first action, correctly predicting both branches remains near
chance. Budget conditioning is thus more reliable over the evolving
closed-loop trajectory than as an immediate switch at $t=0$.

\noindent\textbf{Failure analysis.}
A substantial expert--student gap remains, especially at long horizons. At
$B=800$ and $B=2000$, 92--93\% of the VLN's forward actions revisit traversed
cells. Offline imitation trains on expert trajectories, providing limited
coverage of states reached after student deviations. This pattern motivates
better closed-loop recovery during learned search.

\subsection{Real-World Deployment}

\begin{figure}[ht]
\centering
\includegraphics[width=\linewidth]{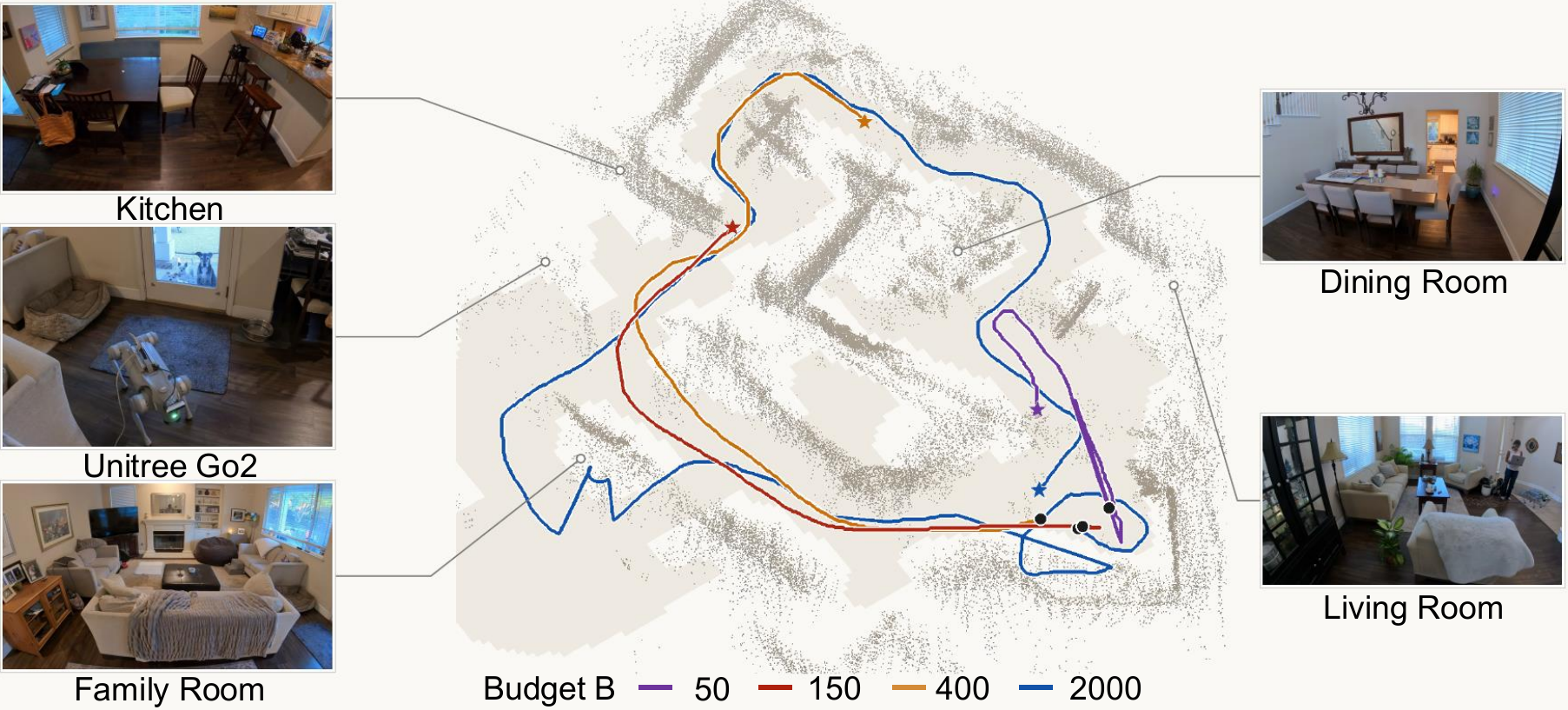}
\caption{\textbf{Real-robot deployment in a multi-room home.}
Routes at $B\in\{50,150,400,2000\}$ share the scene, start, target category,
and weights. At $B=50$, the robot searches for the closest chairs near its
start; with larger budgets, it first makes longer excursions toward the
family room and kitchen. Routes overlay the robot's lidar map; circles mark
starts, stars mark endpoints, and photographs show traversed spaces.}
\label{fig:real}
\end{figure}

We deploy the room-level Bellman-supervised VLN zero-shot on an indoor
quadruped without real-world fine-tuning. Following Uni-NaVid~\cite{uninavid},
an onboard camera streams RGB to a remote GPU server, which returns four
primitives per call in $0.22$\,s on average; locomotion dominates runtime.

As in simulation, forward and turning actions command $0.25$\,m translation
and $30^\circ$ rotation. Odometry executes each primitive until its measured
displacement reaches the target, maintaining consistency with predictions.

In a multi-room home containing 14 chairs, we fix the start pose, target, and
weights and vary only $B\in\{50,150,400,2000\}$ (Fig.~\ref{fig:real}). We run
each budget five times; Table~\ref{tab:real_robot} reports the mean and standard
deviation (SD) of chairs found. We define the success rate in each run as
$\mathrm{SR}=100\times(\text{chairs found}/14)$ and report its mean.

\begin{table}[ht]
\centering
\caption{Real-robot chair search over five runs per budget.}
\label{tab:real_robot}
\footnotesize
\setlength{\tabcolsep}{2.5pt}
\begin{tabular}{lcccc}
\hline
Budget $B$ & 50 & 150 & 400 & 2000 \\
\hline
Chairs found & $3.4{\pm}1.14$ & $5.0{\pm}1.00$ & $7.6{\pm}1.52$ & $9.2{\pm}1.30$ \\
Mean SR (\%) & 24.3 & 35.7 & 54.3 & 65.7 \\
\hline
\end{tabular}
\end{table}

Mean discoveries rise from 3.4 to 9.2 and SR from 24.3\% to 65.7\% as the
budget increases. Routes also differ: at $B=50$, the robot searches near its
start in the living and dining rooms, whereas larger budgets permit longer
transfers to the family room and kitchen. The repeated trials and route changes
show budget-dependent behavior despite physical-camera observations and
locomotion error.

\ifarxiv
\section{Discussion and Limitations}
\label{sec:discussion}

\noindent\textbf{Choosing an abstraction for a student.}
Our results suggest treating the expert's abstraction as a design variable
chosen for the student rather than for expert return. Two requirements pull in
different directions. The abstraction must retain enough structure for
long-horizon, budget-aware planning; without room connectivity and semantic
priors, the expert degenerates toward coverage. It should also keep the
expert's decisions from depending on information that the student can never
recover: revealing target locations makes the expert stronger but, by
Eq.~\eqref{eq:implicit_policy}, asks the student to imitate decisions it
cannot ground in its inputs. Rooms and doorways meet both requirements
reasonably well for indoor search: they are visible in RGB, their semantics
can be learned as cross-scene priors, and their unseen contents must be
discovered by exploring. A heuristic that follows is to retain what the
student can perceive or learn from data, and to withhold what it can only
discover by acting. Whether this heuristic predicts student performance across
tasks and abstractions remains open.

\noindent\textbf{Disentangling granularity and information access.}
The object-level and room-level experts differ in decision unit (instances
versus rooms), in information access (revealed versus withheld targets), and
in execution behavior. Budget-blind controls separate countdown annotations
from the supervision source, but not these factors from each other. Two
additional experts would isolate information access: a room-level expert given
the realized contents of every room, and an object-level expert restricted to
targets it has already observed. Comparing their students with ours would
directly test the information-alignment explanation.

\noindent\textbf{Residual mismatch and memory.}
Even the room-level expert reads room types and objects from simulator
annotations and maintains an explicit room graph, which the student must infer
from RGB history and retain in context. The large remaining expert--student
gap at long horizons, together with frequent revisiting, suggests that spatial
memory rather than supervision limits the student there. Abstractions that the
student can track internally, or students with explicit memory, are natural
next steps.
\fi

\section{Conclusion}

We study budget-conditioned VLN exploration and the transfer of different
expert supervision schemes to Uni-NaVid. Budget-conditioned object search
counts distinct geometric discoveries for a denser evaluation signal than binary success;
controlled countdown interventions show that changing only the stated budget
changes trajectories, beyond extending a fixed route.

\ifarxiv
Framing experts as Bellman planners over abstractions of the full environment
makes the information they act on an explicit design choice: the same
budgeted recursion yields fully informed routing or search under uncertainty,
depending on what the abstraction reveals.
\fi

Expert return and supervision quality diverge: despite higher return from the
fully informed object-level Bellman expert, room-level supervision achieves
the highest mean collection among evaluated VLNs at every budget.
Greedy-frontier supervision also beats object-level supervision at the longest
budget, reinforcing that expert rankings need not transfer. The matched
Bellman comparison supports room-level search as an effective demonstration
scheme for the evaluated VLN; information access, abstraction, and execution
jointly define that scheme.
Budget-blind controls show that room-level supervision retains a mean
advantage without countdown annotations during training, while including
these annotations improves both students at every tested budget.

For real-world demonstrations, these findings motivate recording budget with
observations and actions and considering search under uncertainty. Such
metadata both explains choices between nearby targets and other rooms and
provides a learning signal; whether it improves real-world learning requires
further evaluation.

At 2000 steps, the room-level-supervised VLN remains below its expert
(72.7\% versus 98.0\%). Frequent revisiting and the non-learned frontier
expert's long-horizon advantage motivate better spatial memory and recovery
from trajectory deviations. Rooms are an empirical abstraction, and evidence
is limited to indoor search with equal-value targets and action-count budgets.
Weighted target rewards and nonuniform action costs extend the formulation,
but supervision rankings there remain unknown. Future work can evaluate these settings and
select abstractions for downstream student performance.

\section*{ACKNOWLEDGMENT}

This work used DeltaAI at the National Center for Supercomputing
Applications (NCSA) through allocation CIS261244 from the Advanced
Cyberinfrastructure Coordination
Ecosystem: Services \& Support (ACCESS) program, which is supported by U.S.
National Science Foundation grants \#2138259, \#2138286, \#2138307,
\#2137603, and \#2138296.
We thank Mason Lisi for generously providing access
to their house for real-world robotic experiments.
We also thank Tyler Bonnen for thoughtful discussions.

\bibliographystyle{IEEEtran}
\bibliography{refs}

\end{document}